\documentclass[a4paper, 10pt, conference]{ieeeconf}      % Use this line for a4
\usepackage{FG2021}

\FGfinalcopy % *** Uncomment this line for the final submission
\usepackage{graphicx}
\usepackage[font=small]{caption}
\usepackage{subcaption}
\usepackage{booktabs}
\usepackage{bm}
\usepackage{amsmath,amssymb}
\usepackage{hyperref}

\usepackage{amsthm}
\usepackage{bbm}
\usepackage{cite}
\usepackage[accsupp]{axessibility} 
\hypersetup{
    colorlinks,
    linkcolor={red!50!black},
    citecolor={blue!50!black},
    urlcolor={blue!80!black}
}
\IEEEoverridecommandlockouts      

\def\FGPaperID{155} % *** Enter the FG2021 Paper ID here

\title{\LARGE \bf
Geodesic squared exponential kernel for non-rigid shape registration
}

\author{\parbox{16cm}{\centering
    {\large Florent Jousse $^{1,2}$, Xavier Pennec $^1$,  Hervé Delingette $^1$, Matilde Gonzalez $^2$}\\
    {\normalsize
    $^1$ Université Côte d'Azur, INRIA, EPIONE team, Sophia-Antipolis, France. \\
    $^2$ Qc Labs Department, QuantifiCare, Sophia-Antipolis, France}}
}

\newcommand{\bvec}[1]{\bm{#1}}
\newcommand{\bmat}[1]{\bm{#1}}
\DeclareMathOperator*{\argmin}{arg\,min}

\begin{document}

\IEEEoverridecommandlockouts\pubid{\makebox[\columnwidth]{978-1-6654-3176-7/21/\$31.00~\copyright{}2021 IEEE \hfill}
\hspace{\columnsep}\makebox[\columnwidth]{ }}

\ifFGfinal
\thispagestyle{empty}
\pagestyle{empty}
\else
\author{Anonymous FG2021 submission\\ Paper ID \FGPaperID \\}
\pagestyle{plain}
\fi
\maketitle

%%%%%%%%%%%%%%%%%%%%%%%%%%%%%%%%%%%%%%%%%%%%%%%%%%%%%%%%%%%%%%%%%%%%%%%%%%%%%%%%
\begin{abstract}

This work addresses the problem of non-rigid registration of 3D scans, which is at the core of shape modeling techniques. Firstly, we propose a new kernel based on geodesic distances for the Gaussian Process Morphable Models (GPMMs) framework. The use of geodesic distances into the kernel makes it more adapted to the topological and geometric characteristics of the surface and leads to more realistic deformations around holes and curved areas. Since the kernel possesses hyperparameters we have optimized them for the task of face registration on the FaceWarehouse dataset. We show that the Geodesic squared exponential  kernel performs significantly better than state of the art kernels for the task of face registration on all the 20 expressions of the FaceWarehouse dataset. 
Secondly, we propose a modification of the loss function used in the non-rigid ICP registration algorithm, that allows to weight the correspondences according to the confidence given to them. As a use case, we show that we can make the registration more robust to outliers in the 3D scans, such as non-skin parts. 

%Secondly, we propose to combine NICP with a skin detection algorithm to make it %more robust to reconstruction noise. To do so we modified the registration loss %function to take into account weights on the point pairs. We shows that this %modification avoids the template to fit unwanted areas such as hair and clothes.   

%These instructions provide basic guidelines for preparing camera-ready (CR)
%Proceedings-style papers. This document is itself an example of the
%desired layout for CR papers (inclusive of this abstract). The document
%contains information regarding desktop publishing format, type sizes, and
%type faces. Style rules are provided that explain how to handle equations,
%units, figures, tables, references, abbreviations, and acronyms. Sections
%are also devoted to the preparation of the references and acknowledgments.

\end{abstract}

%%%%%%%%%%%%%%%%%%%%%%%%%%%%%%%%%%%%%%%%%%%%%%%%%%%%%%%%%%%%%%%%%%%%%%%%%%%%%%%%
\section{INTRODUCTION}
    %What is the problem to be solved?
    Shape modeling has applications in various areas such as face synthesis, anthropology and computational anatomy. In this paper, we address the problem of shape registration because many shape modeling methods including the well-established Morphable Models \cite{Blanz1999} require that the shapes are registered (i.e: the shapes are in a comparable space). Moreover, the performances of the shape models are directly related to the quality of the shape registration. A recent survey on the Morphable Models topic and their construction can be found in \cite{Egger2020}.
    
    The methodology to solve the registration problem depends on the nature of the shapes that we observe. When the shapes are 3D scans, represented with meshes, the prevailing approach \cite{Gerig2018,yang2020facescape,Cao2014,FLAME:SiggraphAsia2017} is to first fit a reference mesh, commonly referred as the template, towards the 3D scans. 
    
    Then, to replace the 3D scanned meshes by the deformed templates. 
    In this way, we make sure that all the 3D scans are sharing the same mesh connectivity, which allows to make statistics on the positions of the vertices. 
    
    The standard approach to fit a template mesh 
    is derived from 
    the ICP (Iterative Closest Point) algorithm, whose main idea is 
    to %that we can 
    fit a point cloud to another by minimizing the distance between corresponding points. The algorithm alternates between a transformation of the reference point cloud (that reduces the distance) and an update of the point correspondences.

    Many % Some 
    variations of ICP have been proposed to extend the algorithm to meshes and to non-rigid deformations. Early versions \cite{Besl1992,Zhang1994,Yang1992,Menq1992,Granger2002,Champleboux1992} allowed only rigid motions between shapes, which was not sufficient for shape modeling purposes. Therefore, Feldmard et al. \cite{Feldmar1996} proposed to combine affine and locally affine transformations to allow non-rigid motions, with promising results on a wide range of shapes: teeth, faces, skulls, brains, and hearts. %Thereafter, 
    Later,
    Amberg et al. \cite{Amberg2007} 
    added a stiffness term to the loss function being optimized that penalises differences between the locally affine transformation matrices assigned to neighbouring vertices.

    To make the registration more flexible, Luethi et al. \cite{Luthi2018,Gerig2018} proposed to use Gaussian processes to model ICP shape deformations.
    %allowed into ICP. 
    The framework, coined Gaussian Process Morphable Models (GPMMs) makes use of kernels to add priors on the allowed deformations. Thus, one can make the algorithm specific to the characteristics of the surface being registered. This could allow us to use a specific kernel for different types of surfaces (solids, soft-tissues, bones, etc.) and possibly improve the registration. This method has been used for example to build the Basel Face Model \cite{Gerig2018} or the FaceScape model \cite{yang2020facescape}. 
    
    The main limitation of the GPMMs is that they may fail when the topology of the 3D scans changes too much. This is for example the case for the registration of faces with different facial expressions. Our hypothesis is that this is caused by the kernels that have been proposed, which are all based on the Euclidean distance between the vertices. However, the template being deformed is not a flat surface, which causes the Euclidean distance to be a bad approximation of the true distance along the surface. Consequently, we propose to use geodesic distances into the kernels to make them more specific to the template's geometrical and topological characteristics. 

    An other challenge with non-rigid ICP (NICP) methods is to avoid any over-fitting of the template. Indeed, the closest point heuristic guiding the pairing of points is not always valid and the template mesh may easily fit unnaturally on the target mesh. A common practice is to remove the point correspondences according to specific criteria to make the registration more robust. Classic filtering criteria are based on geometric conditions such as removing the 
    point pairs whose distances are too large.

    Instead of removing 
    the point correspondences
    % matches,
    %pairs, 
    we propose to assign weights to 
    them
    %the correspondences 
    %pairs 
    to allow for more flexible adjustments. For this purpose, we have modified the loss function of the NICP algorithm to incorporate the weights and we describe how the solution can be computed analytically and efficiently. As an example, we show that we can use these weights in combination with a skin detection algorithm to avoid the template to fit to clothes and hair. 

    \section{Organization of the paper}
    In section \ref{sec:relatedWork}, we describe prior work on Gaussian Process Morphable models, non-rigid ICP, and B-spline kernels. 
    
    In section \ref{sec:method}, we introduce the geodesic squared exponential kernel and its computation. We describe how the kernel hyperparameters can be optimized for a specific task. We introduce the weighted loss function. Finally, we describe the experiments done on the FaceWarehouse dataset to optimize and test the kernel.
    
    In section \ref{sec:resutls}, we show some registration results achieved with the different kernels. Then, we show a use case of the weighted loss function. 
    Results are finally discussed in section \ref{sec:discussion}.

\section{RELATED WORK}\label{sec:relatedWork}
%Answer the question "How did you do ? "
%

\subsection{Gaussian Process Morphable Models (GPMMs)}\label{sec:GPMMs}

The key idea of Morphable Models is to model the displacements of the $N$ template vertices $\bvec{U}\in \mathbb{R}^{3N}$ around their mean positions as a joint normal distribution: $\bvec{U} \sim \mathcal{N}(\bvec{\mu},\bvec{\Sigma})$. Note that $\bvec{U}$ has size $3N$ because we model each dimension in a separate random variable. This formulation allows us to construct new shapes by sampling displacements vectors from the distributions with the following steps:
\begin{enumerate}
    \item Find the matrix $\bmat{A}$ such that $\bmat{\Sigma} = \bmat{AA}^{\intercal}$. This can be found through the spectral decomposition $\bmat{\Sigma} = \bmat{Q \Delta Q}^{-1}$ of $\bmat{\Sigma}$ and then $\bmat{A} = \bmat{Q\Delta}^{\frac{1}{2}}$.
    \item Let $\bvec{\alpha} = (\alpha_1, ..., \alpha_{3\times N})^{\intercal}$ be a random vector, $\alpha_i \sim \mathcal{N}(0,1)$.  
    \item Let $\bvec{u} = \bvec{\mu} + \bmat{A}\bvec{\alpha}$ be a sample from the displacement distribution. 
\end{enumerate}

With GPMMs \cite{Luthi2018}, we define a distribution over functions using a Gaussian Process: $u(x) \sim \mathcal{GP}(\mu(x),k(x_i,x_j))$, where $\mu : \mathbb{R}^3 \rightarrow \mathbb{R}^3$ is a vector valued function, that returns the average displacement at vertex $\bvec{x}$ and $k : \mathbb{R}^3 \times \mathbb{R}^3 \rightarrow \mathbb{R}^{3\times 3}$ is a matrix valued kernel. 
The Gaussian process is defined by the mean function $\mu(x)$ and the kernel $k(x_i,x_j)$. Therefore, the assumptions about the distribution are encoded into the kernel, which allows for example to integrate spatial correlations between neighboring displacements. We can also sample from the Gaussian Process to generate new shapes with the following steps: 
\begin{enumerate}
    \item Find the eigenvalues/eigenfunctions $\lambda_i$ and $\phi_i$ of the integral operator $\tau_k$ evaluated on the domain $\Omega$ (i.e: the mesh vertices): $\tau_k f(.):= \int_\Omega k(x,.)f(x)d\rho(x).$ More details about the computation of the eigenfunctions using the Nyström method can be found in \cite{Rasmussen2018}. 
    \item Let $\bvec{\alpha} = (\alpha_1, ..., \alpha_{3\times N})^{\intercal}$ be a random vector, $\alpha_i \sim \mathcal{N}(0,1)$. 
    \item \label{step3} Let $u(x)= \mu(x) + \sum_{i = 1}^{\infty} \alpha_i \sqrt{\lambda_i} \phi_i(x)$ be a sample from the distribution. 
\end{enumerate}

In practice, the infinite sum in step \ref{step3} is truncated realizing the following low-rank approximation: $u(x)= \mu(x) + \sum_{i = 1}^{\infty} \alpha_i \sqrt{\lambda_i} \phi_i(x) \approx \sum_{i = 1}^{r} \alpha_i \sqrt{\lambda_i} \phi_i(x)$. The low rank approximation works well when the eigenvalues values are decreasing rapidly since the approximation error is equal to the tail of the sum:$\sum_{i=r+1}^\infty \lambda_i.$
This formulation provides a model of vertex displacement parameterized by the $\alpha_i \in \mathbb{R}$, which can be used to deform the template mesh.

We can rewrite this parametric model in matrix form
\begin{equation}\label{eq:matrixModel}
    \bvec{y} = \bmat{X} \bvec{\alpha},
\end{equation}
where $\bmat{X}$ is the matrix containing the evaluation of the eigenfunctions at mesh vertices:
\begin{equation}
    \bmat{X}= 
    \begin{pmatrix}
    \sqrt{\lambda_1}\phi_1(x_1)  && \hdots && \sqrt{\lambda_r}\phi_r(x_1) \\
    \vdots && \ddots && \vdots \\
    \sqrt{\lambda_1}\phi_1(x_N) && \hdots && \sqrt{\lambda_r}\phi_r(x_N) 
    \end{pmatrix},
\end{equation}
and $\bvec{\alpha} = (\alpha_1 \hdots \alpha_r)^\intercal$ is the vector of parameters.

\subsection{B-spline kernels}
% Il manque des phrases de transitions
We have seen that GPMMs allow to use a prior for the distribution of vertices displacements around the mean shape. In the field of face registration, Gerig et al. \cite{Gerig2018} proposed to use the B-Spline kernel of Opfer et al.\cite{Opfer2006}, which % can be written as 
is
a sum over the B-spline support parameterized by $\sigma$:
\begin{equation}
    k_{Sp}(x_i,x_j,\sigma) = \sum_{k\in\mathbb{Z}^d} 2^{2-\sigma}\zeta(2^\sigma x_i-k)\zeta(2^\sigma x_j -k).
\end{equation}
The function $\zeta$ is built with third order univariate B-spline as: $\zeta(x) =b_3(x_1)b_3(x_2)b_3(x_3)$ and the hyperparameter $\sigma$ is the spline's support. The univariate spline function $b_3$ is a function with compact support defined as: 
\begin{equation}
    b_3(x) = 
     \begin{cases}
       \frac{2}{3}-|x|^2 +\frac{1}{2}|x|^3 &\quad\text{if} \quad 0\leq|x|<1,\\
        \frac{(2-|x|)^3}{6}     &\quad\text{if}\quad 1\leq|x|<2,\\
       0 &\quad\text{else}.\\
     \end{cases}
\end{equation}

The measure of similarity between the displacement of vertices is controlled by $\sigma \in \mathbb{R}$, such that increasing $\sigma$ results in smoother deformations and vice versa. Consequently, the kernel value is related to the support of the spline, which means that vertices that are close in Euclidean space $\mathbb{R}^3$ must follow similar deformations. Because of this property, the B-spline kernel do not allow the mouth or the eyes to open and close properly, which is important to register facial expressions.

To increase the expressiveness of the kernel, Gerig et al.\cite{Gerig2018}  proposed to sum several B-splines kernel with different values of hyperparameter $\sigma_l\in \mathbb{R}$ and weight $s_l\in \mathbb{R}$:
\begin{equation}
    k_{BSp}(x_i,x_j)= I_{3\times3}\sum_{l\in L}s_l k_{Sp}(x_i,x_j,\sigma_l).
\end{equation}
The authors proposed to use 4 levels but to our knowledge the weights have not been made public. We  study the impact of multiscale versus single scale B-spline kernels in \ref{sub:scoreplots}.

Despite its multiscale property, this kernel may still not be used for shape registration when the topology of the meshes changes too much either 
% such as 
%For example, for registering facial expressions 
because the similarity between the vertices is still a function of the Euclidean distance.

A solution proposed by Gerig et al. In \cite{Gerig2018} to register facial expressions consist in using a statistical prior instead of the kernel. This makes the task of building the face Morphable model more difficult to implement because building a statistical prior for a morphable model requires a registered mesh dataset (e.g. PCA based morphable models in \cite{Blanz1999}).

\subsection{Non-rigid ICP}
 
The GPMM model described previously can be used in combination with NICP for shape registration \cite{Gerig2018}. In this case, NICP searches the set of parameters $\bvec{\alpha}$ minimizing the distance between a reference mesh $V$ and a target mesh $W$.
The performances of NICP depends on the initial alignment of the meshes. Thus, a set of landmarks can be used to guide the alignment of the template mesh to the 3D scanned mesh. Three landmarks points are sufficient to retrieve a similarity transform (translation, rotation and uniform scaling) that moves the template mesh towards the target. 

Also, if we know the ideal positioning of some landmarks points we can combine this information with our prior model into a posterior model. For a set of $H$ observed landmarks, $\bvec{y}\in\mathbb{R}^{3H}$ is the column vector containing the displacements of the landmarks between the template and the target. Let $k_*(x)$ be the vector containing the kernel prior values between the test point $x$ and the $H$ training points. Using theses notations we can write the posterior equations: 
\begin{align}
    u_*(x) &= k_*(x)^{\intercal}[\bmat{K}+\sigma_n^2I]^{-1}\bvec{y},\\
    %\begin{split}
         k_*(x_i,x_j) &= k(x_i,x_j) %\\
     %&
     -k_*(x_i)^{\intercal}[{\bmat{K}}+\sigma_n^2I]^{-1}k_*(x_j).
    %\end{split}
\end{align} This posterior model can be used in place of the prior model to perform the registration of the template.

In addition to the initialization and the posterior computation, each iteration of NICP contains the following steps: firstly the reference mesh is fitted to the target mesh by minimizing a loss function, that includes the distance between the point correspondences and a regularization term. Then, the correspondences are updated by taking the closest points. Point correspondences are then fileterd to keep only those whose normals are close and are symmetric  $(a,b)$ (i.e: if $a \in V$ is the closest point from $b\in W$, then $b\in W$ has to be the closest point from $a \in V$). These steps are iterated with a decreasing regularization rate to achieve a coarse-to-fine registration.  

The described procedure works fine if the 3D scans are acquired in a controlled environment but may fail in certain conditions. For example, 
a bald face template may fit the hairs of a hairy scan without constraints in face registration,
%for face registration if any hair is present on the 3D scans the template will fit without constraints,
which is not satisfactory. We propose some modifications on the loss function in \ref{sub:Confidence} to address this issue. %To make the procedure robust to such cases we implemented a skin detection algorithm and modified the loss function of NICP to take into account weights on the point pairs. The description of the method is provided in section \ref{sub:Confidence}. 

%\subsection{Geodesic Gaussian processes}

%%In the field of surface reconstruction, Del Castillo et al.% \cite{DelCastillo2015} proposed to use a Gaussian process to infer the% surface that passes through a point cloud. They shown that using% geodesics into the kernel yields to better results than using the% Euclidean distance.  To do so, they propose to first search an% isoparameterisation $p: S \in \mathbb{R}^3 -> T \in \mathbb{R}^2$ of% the point cloud and then compute the geodesics as "straight lines" in% the parameterization. Then the surface reconstruction problem is posed% as a kriging problem. The limitations that apply to surface% parameterization algorithms are also valid for this method. For example% the algorithm may have troubles with severe curvature areas and sharp% edges. Besides, the isoparameterization does not always exists (for% example for a sphere mesh) while the geodesics does.
%
%The use of geodesic exponential kernels in a Gaussian process has %remained limited and to our knowledge has not been used for shape %registration. 

\section{METHODS}\label{sec:method}

\subsection{Geodesic squared exponential kernel}

As we have seen, the main problem with the kernels proposed with GPMMs (B-spline kernels or Euclidean squared exponential kernel) is that the behavior of vertices displacements is a function of the Euclidean distance. Thus vertices at equal distance in $\mathbb{R}^3$ behave similarly without any concern about the specificity of the mesh surface. 

\begin{figure}[!t]
\centering
\begin{subfigure}{.249\textwidth}
  \centering
  \includegraphics[width=1\textwidth]{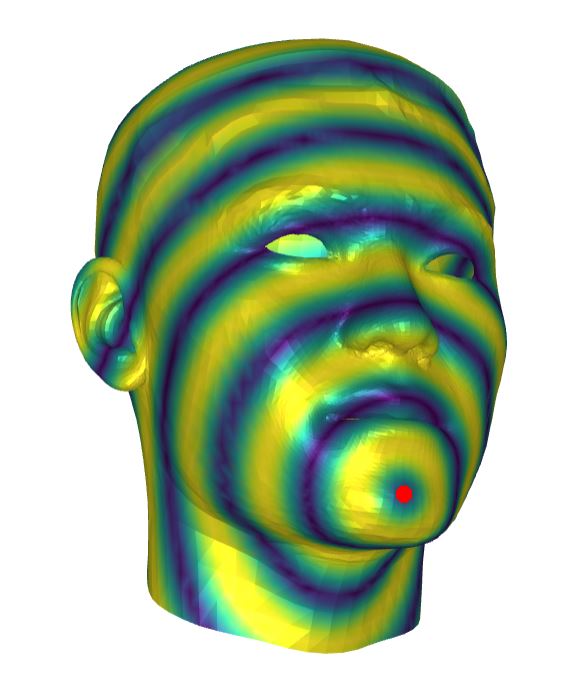}
  \caption{Euclidean distance }
  % \label{fig:sub1}
\end{subfigure}%
\begin{subfigure}{.249\textwidth}
  \centering
  \includegraphics[width=1\textwidth]{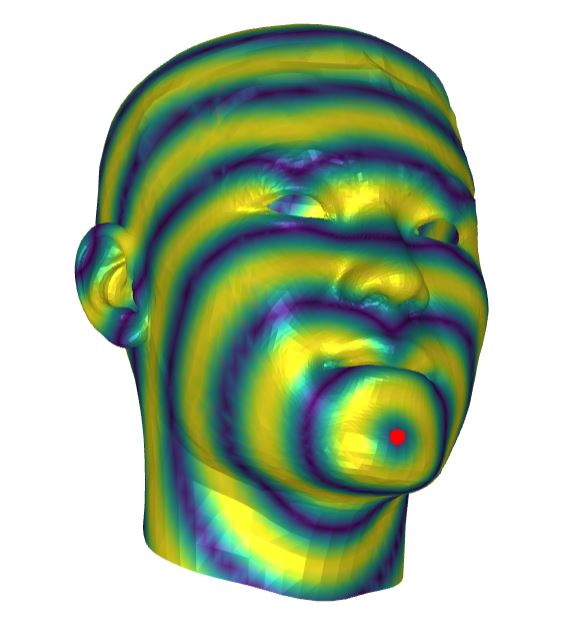}
  \caption{Geodesic distance }
  % \label{fig:sub2}
\end{subfigure}%
%ABl: J'ai essayé de reformuler la légende pour aller plus loin qu'une description sommaire
\caption{Isolines of distances from a point of a mesh to the whole mesh for two metrics (Euclidean and Geodesic). The points that are on the same isoline are at equal distance from the red point. This highlights that holes in the mesh (eg. mouth and eyes) are only correctly taken into account by the geodesic metric, effectively increasing the distance between parts which are geometrically close but not from a topological point of view. In the context of GPMMS, this means that such parts will have a lower correlation, allowing for more degrees of freedom during the deformation.}
\label{fig:distanceComparison}
\end{figure}

To solve this issue we propose another kernel based on geodesic distances (i.e.: the shortest distance along the mesh surface). It can be used to improve the results of facial expression registration because it allows mouth openings and closings. In a more general context, this kernel will produce more realistic deformations around holes and curved areas on the template. As we can see from Figure \ref{fig:distanceComparison}, using geodesic distances reduce the similarity between vertices separated by holes or in high curvature areas. For example, vertices around the mouth are more distant with the geodesic distance than with the Euclidean distance. Thus, the kernel becomes more specific to the surface. We will show that this leads to better face registration and more realistic deformations in the results section \ref{sec:resutls}.

Similarly to the Euclidean squared exponential kernel (Euclidean SE kernel), we can write the geodesic squared exponential kernel (geodesic SE kernel):
\begin{equation}
    k_{se}(x_i, x_j) = I_{3\times3}\cdot s \cdot exp( - \frac{\gamma(x_i,x_j)^2}{2\sigma^2}),
\end{equation}
with $\gamma$ being the geodesic distance on the mesh. The kernel is parameterized by the length scale $\sigma \in \mathbb{R}$ which determines the shape of the Gaussian (i.e. larger $\sigma$ induce smoother deformations) and a scaling factor $s\in \mathbb{R}$. Note that the support of a B-spline kernel is compact, that is not the case for squared exponential kernels. Furthermore, here the parameter $\sigma$ correspond to the variance and not the support. Thus, we expect to use larger sigma with B-spline kernels than with squared exponential kernels.
%a quoi correspond les valeurs propres nulles (précision numérique 
%
Following the definition of multi-scale kernels in \cite{Gerig2018}, several geodesic kernels can also be summed together with varying $\sigma$ to build a multi-scale kernel.

In the field of surface reconstruction, Del Castillo et al. \cite{DelCastillo2015} propose to use a Gaussian process to infer the surface that passes through a point cloud. They have shown that using geodesic distances into the kernel yields to better results than using the Euclidean distance. They propose to first search an isoparameterisation $p: S \in \mathbb{R}^3 \longrightarrow T \in \mathbb{R}^2$ of the point cloud and then compute the geodesic distances as "straight lines" in the new parameterization. Then the surface reconstruction problem is posed as a kriging problem. The limitations that apply to surface parameterization algorithms are also valid for this method. For example, the algorithm may have troubles with severe curvature areas and sharp edges. Besides, the isoparameterization does not always exists (for example for a sphere mesh) while the geodesic distance does. Apart from this, the use of geodesic SE kernels in a Gaussian process has remained limited and to our knowledge it has not been used for shape registration.

\subsection{Geodesic distance computation}

The geodesic distance can be computed efficiently via the heat method proposed in \cite{Crane2013}. The method involves only the resolution of two partial differential equations, which is computationally efficient in practice: 
\begin{enumerate}
    \item The first step is to integrate the heat equation for some fixed $t$. It can be compactly written with the Laplacian:
        $
         \frac{\delta u}{\delta t} = \Delta u,
       $
        where $u$ is the temperature after time $t$. 
    \item The second step is to evaluate %% $\bmat{X}$ %% why bold when it is not bold later?
    the normalized gradient of $u$: 
    $
        X =  -\frac{\nabla u}{|\nabla u|}.
    $
    \item Finally the Poisson equation is solved to retrieve the geodesic distance:
    $
        \Delta \gamma = \nabla \cdot X.
    $
\end{enumerate}

Practically, the heat method works on any surface discretization with proper definitions of gradient $(\nabla)$, divergence $(\nabla \cdot)$ and Laplacian $(\Delta)$ (see \cite{Crane2013} for definitions and computation on simplicial meshes). The geodesic SE kernel can be evaluated on the same domain as the geodesic distances. 

Notice that the computation time of geodesic distances is not a limitation because the geodesic distances can be computed once and reused for further registration with the same template.

\subsection{Positive definiteness of geodesic exponential kernels}

Many properties, such as Mercer's theorem, are based on the positivity of the kernel. 
The geodesic SE kernel that we proposed to use is a particular case of the kernels studied in Feragen et al. \cite{Feragen2015} referred as the geodesic Gaussian kernel (i.e. in our case $q=2$ in equation (1) of \cite{Feragen2015}).
For this particular case, it has been shown that the kernel is positive definite if and only if the space is flat for all $\sigma > 0$. 
Non-positive kernels can produce negative eigenvalues, whose square root leads to complex values in the Karhunen-Loève decomposition of the GP (see the sampling from a GP in section \ref{sec:GPMMs}). However, negative eigenvalues can be discarded by setting an appropriate $r$. In practice, no negative eigenvalues have been encountered in our experiments with $r<1000$ and $ 10^0 < \sigma < 10^4$.

In NICP, the predictive equations for Gaussian process regression are used to combine the prior kernel with the known displacements between corresponding landmarks. These equations requires the inversion of the Gram matrix (see \cite{Rasmussen2018} chapter 2). Therefore, in the general case the kernel must have non-zero eigenvalues. 
But in practice, the observations (i.e. displacements between landmarks) are always noisy. Taking that noise into account in the predictive equations involves adding a constant to the diagonal of the Gram matrix. This addition ensures that we have non-null eigenvalues and thus that the matrix is invertible.

\subsection{Tuning of kernel's hyperparameters for face registration}\label{sub:tuningKernel}

We propose to compare the kernels (geodesic SE, Euclidean SE and B-Spline) by observing the energy, that remains after fitting the GPMM to a given mesh. For this purpose, we use the FaceWarehouse dataset \cite{Cao2014} as a ground truth, which contains 3000 3D face meshes with facial expressions that are already registered. 
We assume that the point correspondences are correct. 
%We consider the pairing of points correct.

Fitting the GPMM to the $i$-$th$ mesh is a linear regression problem where we try to minimize the residuals $\epsilon$ in the following model:
\begin{equation}
    \bvec{y_i}  = \bmat{X} \bmat{\alpha} + \epsilon.
\end{equation}
Here $\bvec{y_i} = \bvec{\Bar{V}} - \bvec{V_i} \in \mathbb{R}^{3N}$ is the vector containing the displacements of the vertices from the average mesh. The matrix $\bmat{X} \in \mathbb{R}^{3N\times r}$ is the model containing the basis vectors of the GPMMs (see equation \eqref{eq:matrixModel}), that is dependent on the choice of kernel and associated hyperparameters. The number of regressors $r \in \mathbb{N}$ is determined by the size of the low rank approximation described in section \ref{sec:GPMMs}. For the experiments, the bounds of the sum are fixed to $r = 1000$ to keep a reasonable computation time. 

For a given mesh and kernel we optimize the coefficient vector $\alpha$
%% XP: on optimise alpha et le resultats de cette optimisation est \hat \alpha.
%$\hat{\bvec{\alpha}}$ 
by ridge regression. The loss function includes a term to penalize the complexity of the deformation (i.e. penalize large $\alpha$
%$\hat{\alpha}$ 
values): 
\begin{equation}\label{eq:loss}
    \hat{\bvec{\alpha}} = \argmin_{\alpha} \frac{1}{2} || \bvec{y} - \bvec{X} \bvec{\alpha} ||^2 + \frac{\rho}{2}||\bvec{\alpha}||^2.
 \end{equation}
Optimal parameters are found when the gradient vanishes:
%The parameters can then be found by setting the gradient to zero: 
\begin{equation}
    \hat{\bvec{\alpha}} = (\bmat{X^{\intercal} X} + \rho \bmat{I}_r)^{-1} \bmat{X}^\intercal \bvec{y}.
\end{equation}

For a given kernel hyperparameter $\sigma$ and template to target displacement $\bvec{y_i}$ the regression energy $R$ is the loss function described in equation \eqref{eq:loss} evaluated at $\hat{\bvec{\alpha}}$: 
\begin{equation}
    R =\frac{1}{2} || \bvec{y} -  \bvec{X} \hat{\bvec{\alpha}} ||^2  + \frac{\rho}{2}||\hat{\bvec{\alpha}}||^2.
\end{equation}

We performed a grid search for each kernel to find the hyperparameters that optimize the energy $R$ for two meshes simultaneously by summing the energy of the two registrations. A mesh with neutral expression and another with open mouth has been chosen, so that the change of expression is important. For the SE kernels, $\sigma$ refers to the lengthscale which controls where is the inflexion point of the Gaussian and for the spline kernels $\sigma$ is the width of the spline support. Thus, in any case the hyperparameter $\sigma$ has an impact on the similarity between vertices displacements. We also compared multiscale kernels by summing several kernels with increasing $\sigma$ (i.e. $\sigma_n = 2 \times \sigma_{n-1}$). The weights of these multiscale kernels are built in the same manner. 

\begin{figure}[tp]
%\centering
\begin{subfigure}{0.48\textwidth}
  %\centering
  \includegraphics[width=1.0\textwidth]{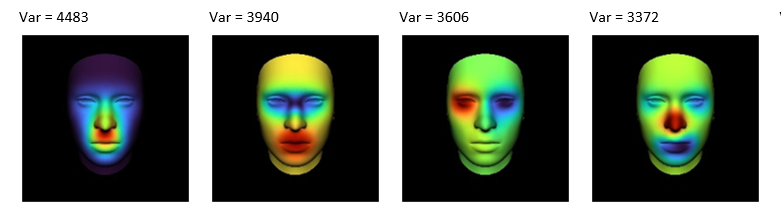}
  \caption{$\sigma = 19$}
  % \label{fig:sub2}
\end{subfigure}

\begin{subfigure}{0.48\textwidth}
  %\centering
  \includegraphics[width=1.0\textwidth]{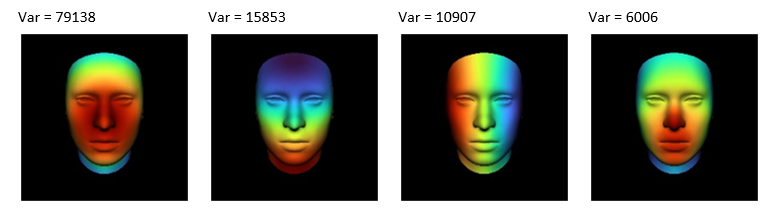}
  \caption{$\sigma = 150$}
  % \label{fig:sub2}
\end{subfigure}

\caption{Visualization of the eigenfunctions of the Euclidean SE kernel in function of the hyperparameter $\sigma$. Lower $\sigma$ leads to more localized eigenfunctions, also the variance associated to the eigenfunction is lower. The color scale shows the displacements such that two vertices with the same color move in the same direction.}
\label{fig:eigenvectors}
\end{figure}

To better understand the effect of the hyperparameters we propose also to vizualize the spectrum of the Gram matrix $\bmat{K} \in \bmat{R}^{N\times N}$, whose entries are $K_{ij}=k(x_i,x_j)$. For this, we compute the eigenvectors of $\bmat{K}$ and then we map the coefficients of the eigenvectors to RGB colors that can be displayed on 3D mesh surface. In this way two vertices with the same displacement will be displayed with the same color. Figure \ref{fig:eigenvectors} shows that reducing $\sigma$ leads to more localized deformations on the template mesh. 

%TODO pas terrible
After the hyperparameter optimization we computed the energy $R$ for each kernel on the whole 3000 samples of the FaceWarehouse dataset. This allow to validate that the performance improvement provided by using the geodesic SE kernel extends to all faces from the dataset and is not specific to the mesh used to learn the hyperparameter. We also computed for each 20 facial expressions (called pose in the dataset) the mean and standard deviation of the energy $R$ over the 150 samples faces. As a remainder, this process is done with the hyperparameters found previously. Then we compared for each facial expression which kernel performs better by looking at the average and standard deviation of the energy $R$ over the 3000 face meshes.

\subsection{Confidence map in NICP} \label{sub:Confidence}

As mentioned earlier, the robustness of NICP can be improved by taking into consideration the quality of the point correspondences. Indeed, there are regions of the 3D scans that are more likely to be noisy (hair for example) and these regions might not be of interest to model the face shape.

We propose to modify the loss function of NICP proposed in \cite{Gerig2018} to take into account a notion of confidence between the point correspondences such that the more uncertain point correspondences have less impact in the minimization. To do so, we add weights $w_j$ to the residuals, which leads to the modified loss function:

\begin{equation}
   \begin{split}
       E(\alpha) &= \frac{1}{2} \sum_{j=1}^{3N} w_j^2 (y_j - \sum_{i=1}^{M}(x_j^i \alpha_i))^2 + \frac{\rho}{2} \sum_{i= 1}^{M}(\alpha_i)^2,\\
         &= \frac{1}{2} ||\bmat{W}( \bmat{Y} - \bmat{X} \bvec{\alpha} )||^2 + \frac{\rho}{2}||\bvec{\alpha}||^2.
   \end{split}
\end{equation}
The parameters $\hat{\bvec{\alpha}}$ minimizing this energy can be found analytically by differentiation: 
\begin{equation}
      \hat{\bvec{\alpha}} = (\bmat{X}^\intercal \bmat{W} \bmat{X} + r \bmat{I}_{M \times M})^{-1} \bmat{X}^\intercal \bmat{W} \bmat{Y}.
\end{equation}
     
In practice, this modified loss function allows to filter the point correspondences based on some criterion. % 
These weights can be used for example to penalize the correspondences that are not on the skin and avoid the template to fit to clothes and hair. For such an application, we computed a confidence map using a skin detection algorithm (\cite{Kolkur2017}). Then during the registration we assign to each point pair a weight $w$ whose value corresponds to the pixel value in the confidence map.

%Metric units are preferred for use in IEEE publications in light of their
%international readership and the inherent convenience of these units in many fields.
%In particular, the use of the International System of Units (SI Units) is advocated.
% This system includes a subsystem the MKSA units, which are based on the
% meter, kilogram, second, and ampere. British units may be used as secondary units
% (in parenthesis). An exception is when British units are used as identifiers in trade,
% such as, 3.5 inch disk drive.

\addtolength{\textheight}{-2.7cm} 
  % This command serves to balance the column lengths
                                  % on the last page of the document manually. It shortens
                                  % the textheight of the last page by a suitable amount.
                                  % This command does not take effect until the next page
                                  % so it should come on the page before the last. Make
                                  % sure that you do not shorten the textheight too much.

%%%%%%%%%%%%%%%%%%%%%%%%%%%%%%%%%%%%%%%%%%%%%%%%%%%%%%%%%%%%%%%%%%%%%%%%%%%%%%%%
\section{RESULTS}\label{sec:resutls}

\subsection{Regression scores plots in function of kernels and hyperparameters}\label{sub:scoreplots}

\begin{figure*}[tp]
%\centering
\begin{subfigure}[t]{0.5\textwidth}
  \centering
  \includegraphics[width=1.0\textwidth]{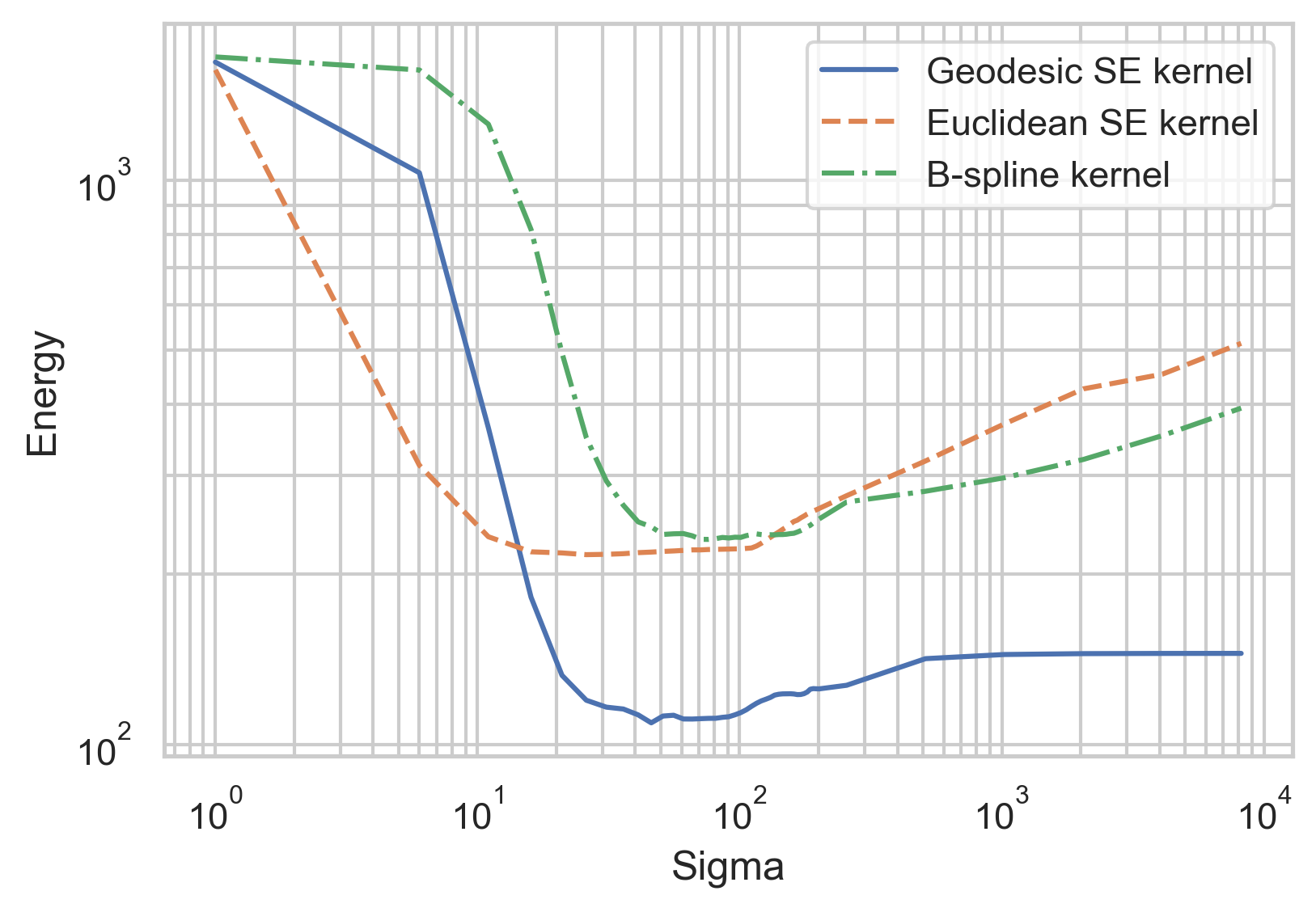}
  \caption{Minimum is reached for $\sigma=70$ with the Spline kernel, $\sigma=25$ with the Euclidean kernel and $\sigma =45$ with the geodesic kernel.}
  % \label{fig:sub1}
\end{subfigure}
\begin{subfigure}[t]{0.5\textwidth}
  %\centering
  \includegraphics[width=1.0\textwidth]{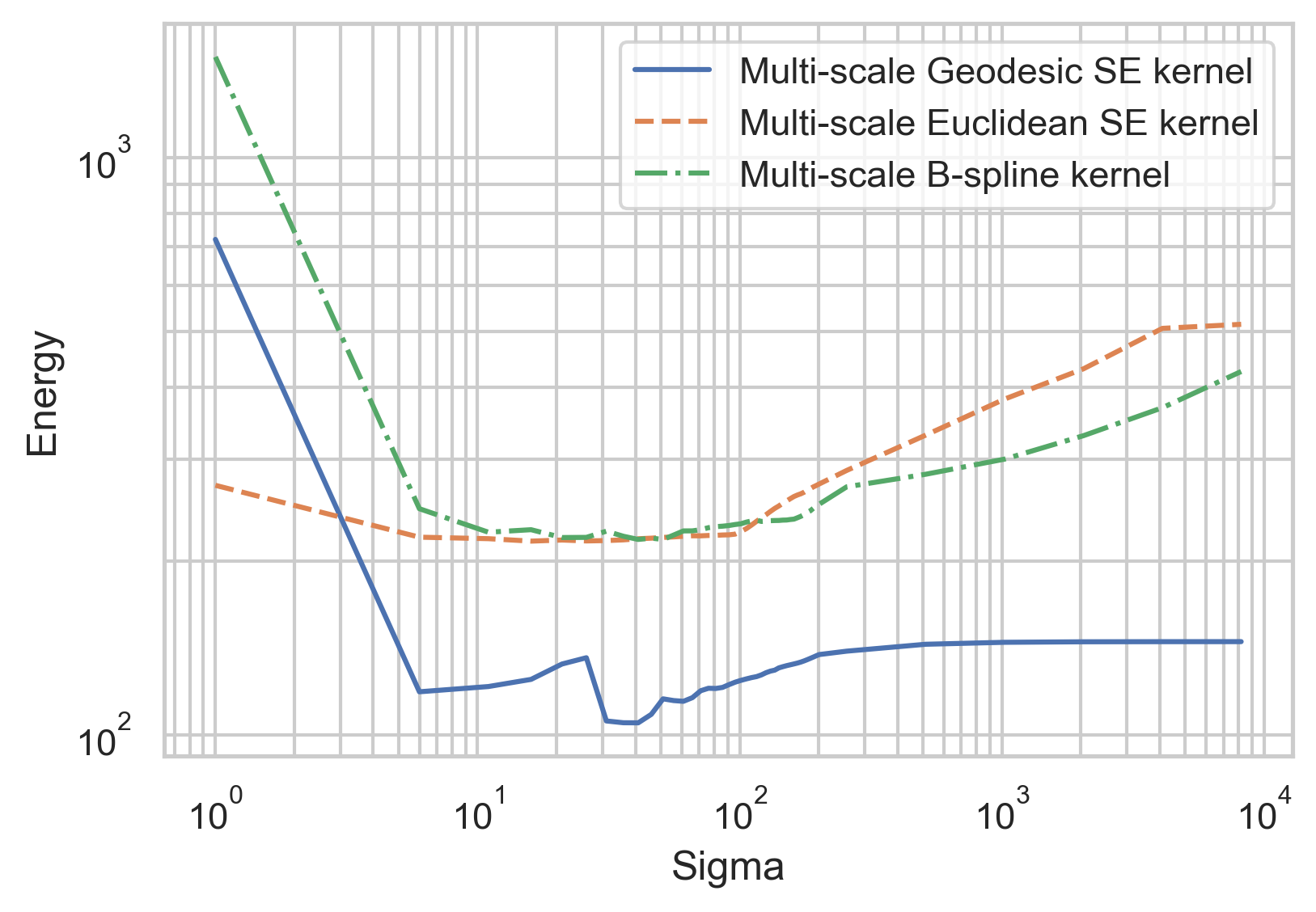}
  \caption{In this plot $\sigma_1$ is the lengthscale of the lower level, 4 kernels are summed with $\sigma$ multiplied by two each time. Minimum is reached for $\sigma_1=40$ with the Spline kernel, $\sigma_1=11$ with the Euclidean kernel and $\sigma_1 =35$ with the geodesic kernel.}
  % \label{fig:sub2}
\end{subfigure}
\caption{These plots compare the energy of registration in function of the hyperparameter sigma for the 3 kernels discussed (with one scale or multiple scales). The energy has been computed for two facial expressions (neutral and open mouth). The geodesic SE kernel has an energy two times lower in both cases which indicates a better fit of the surface.}
\label{fig:regressionScores}
\end{figure*}

Our first experiment has been to optimize the hyperparameters of each described kernel (SE kernels, B-spline kernels and multiscale kernels). Figure \ref{fig:regressionScores} shows the plots of energy $R$ against the hyperparameter $\sigma$ for all kernels. 
%ABl: J'ai l'impression qu'il y a des répétitions (deux fois geodesic SE kernel and deux fois B-spline kernel) mais avec des valeurs différentes
The optimal hyperparameters and their associated energies are: $\sigma=25,  R=216$ for the Euclidean SE kernel ; $\sigma_1=11, R=216$ for the multi scale Euclidean SE kernel ; $\sigma= 45,  R = 109$ for the multi scale  geodesic SE kernel ; $\bm{\sigma_1= 35, R = 104}$ for the  multi scale geodesic SE kernel ; $\sigma= 70, R = 230$ for the B-spline kernel ; and $\sigma_1= 40, R = 217$ for the multi scale  B-spline kernel. 
The energy of regression is not really meaningful if we look at it in an absolute way. However, it is important to see that the use of geodesic distances reduce the regression energy by a factor of two and that other kernels have similar energy values.
The results are discussed in a more thorough way in the section \ref{sec:discussion}.

\subsection{Qualitative comparison of the template registration}

\begin{figure}[tp]
\centering
%\begin{subfigure}{.45\textwidth}
  \centering
  \includegraphics[width=.48\textwidth]{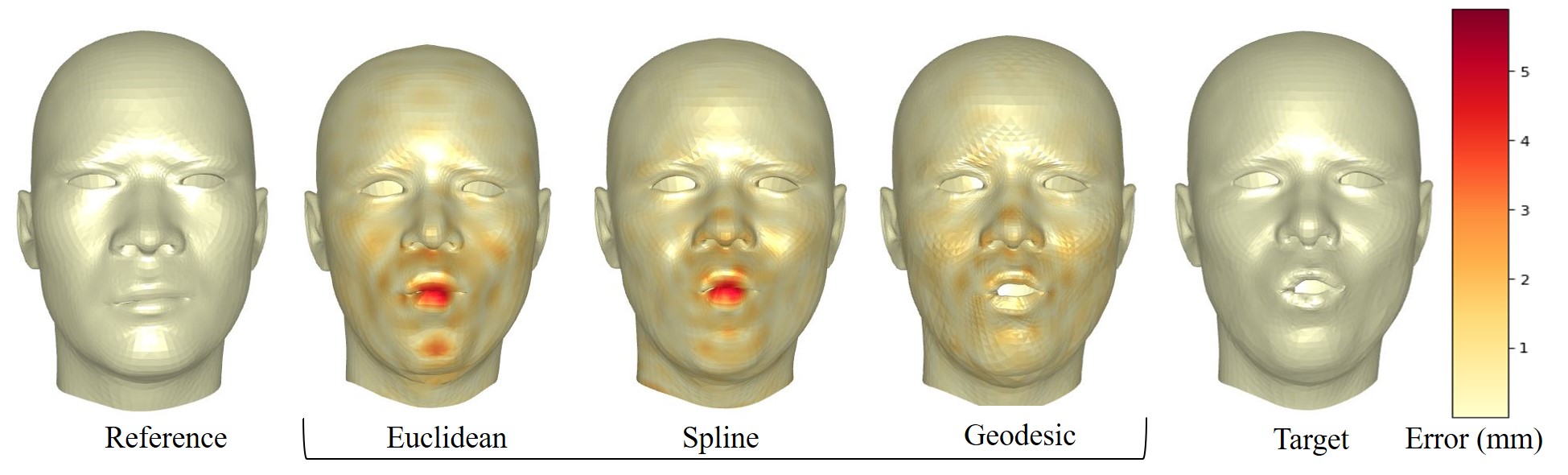}
%  \label{fig:sub1}
%\end{subfigure}%
%\begin{subfigure}{.45\textwidth}
% \centering
  \includegraphics[width=.48\textwidth]{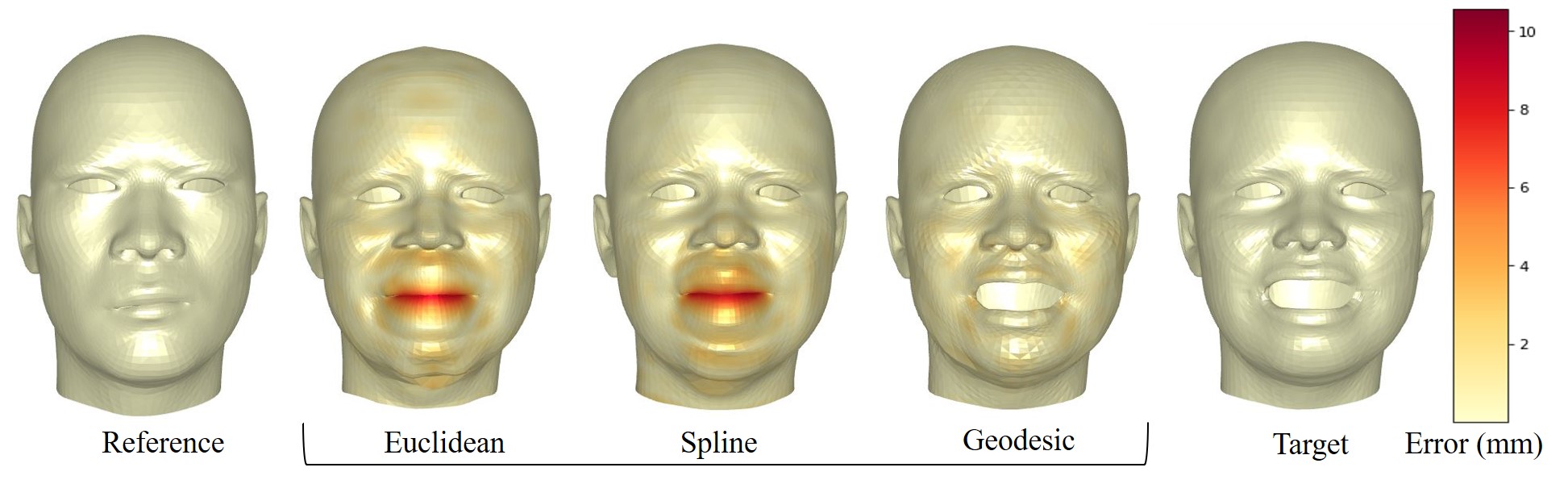}
%  \label{fig:sub2}
%\end{subfigure}%
%\begin{subfigure}{.45\textwidth}
%  \centering
  \includegraphics[width=.48\textwidth]{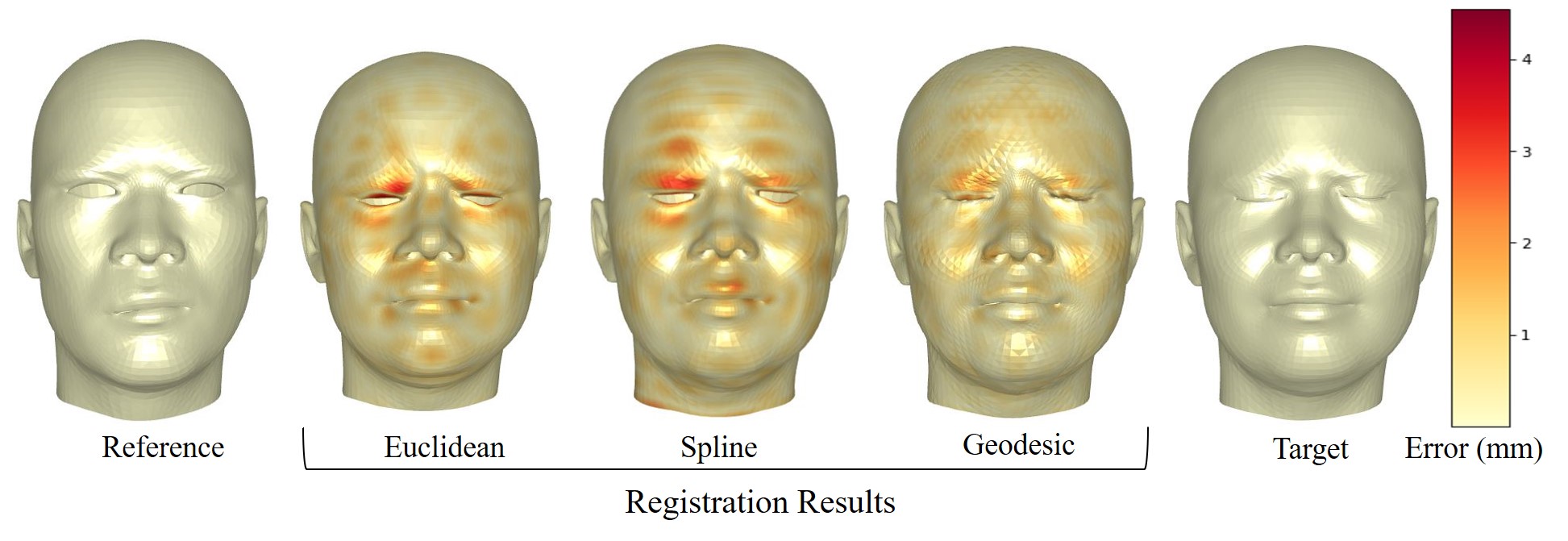}
% \label{fig:sub3}
%\end{subfigure}
\caption{Result of the registration of a template (left column) to a target mesh (right column) for 3 faces with different expressions. The meshes in the central columns are the results of the registration (i.e. the deformed template) that are obtained with the different kernels. It should be noted that using the geodesic SE kernel results in a better fit (lower error) around the eyes and the mouth. B-spline kernel and Euclidean SE kernel gives similar results.}
\label{fig:regressionResults}
\end{figure}

The qualitative registration results obtained with the geodesic SE kernel, Euclidean SE and the B-spline kernel (single scales) are shown on Figure \ref{fig:regressionResults}. The colorscale indicates the error in millimeters. We observe a larger registration error around the mouth and the eyes with the B-spline kernel and the Euclidean SE kernel than with the geodesic SE kernel. The hyperparameters used to generate these results are those obtained using the optimization described in section \ref{sub:tuningKernel}.

\subsection{Comparing regression energy in function of the facial expression} 

Then we compared the registration results on the whole FaceWarehouse dataset. The mean and standard deviation of registration energy for each facial expression are shown in table \ref{tab:resultsPose}. In these results, the kernel hyperparameters are fixed with the values found in previous section. Averaging on all expressions, the Euclidean SE kernel (single scale) obtain a minimum energy $R = 77 (\pm 45)$, the geodesic SE kernel (single scale) obtain a minimum energy $\bm{R = 54 (\pm 5)}$, and the B-spline kernel (single scale) obtain a minimum energy $R = 79 (\pm 45)$. We can see that the standard deviation of regression energy is lower with geodesic SE kernel which means more robustness.
For example, for the face registration with smile expression, we obtain an energy of $R=59$ with the geodesic distance compared to $R=245$ with the Euclidean distance and $R=245$ with the B-spline kernel.

\subsection{Registration results on 3D scan}

\begin{figure}[!t]
% \centering
%\begin{subfigure}{.45\textwidth}
 % \centering
  \includegraphics[width=.48\textwidth]{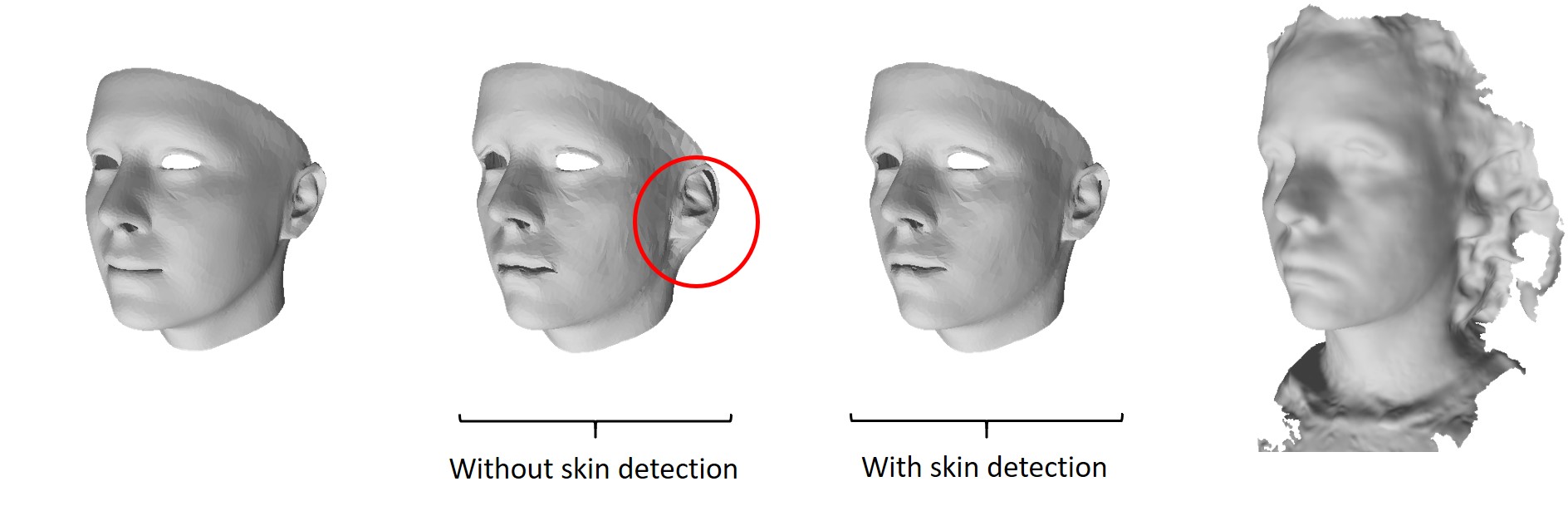}
 %\label{fig:sub1}
%\end{subfigure}%
%\begin{subfigure}{.45\textwidth}
%  \centering
  \includegraphics[width=.48\textwidth]{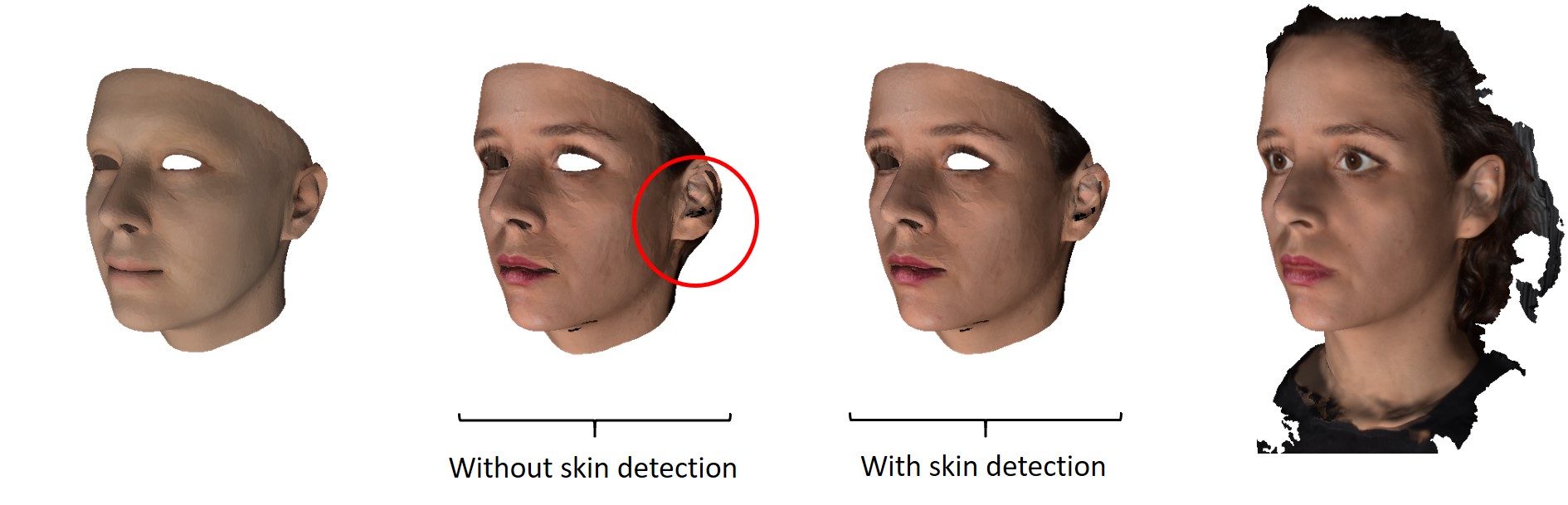}
  %\label{fig:sub2}
%\end{subfigure}%

\caption{This shows the registration result of fitting a template (left column) to a 3D scan (right  column). The meshes in the central columns are the results of the registration (i.e. the deformed template) that are obtained with and without our modified loss function. The loss functions allows to penalize point correspondences that are not on the skin. Without the skin detection we can see that the template fits some hair in the red area. The 3D scans have been acquired with QuantifiCare LifeViz\textregistered~Mini.}
\label{fig:registrationWithTexture}
\end{figure}

A qualitative result of registration of our template on a 3D scan is shown in Figure \ref{fig:registrationWithTexture}. We used as a target a 3D scan acquired with QuantifiCare LifeViz\textregistered~Mini. NICP as it is described in the previous section has been used. The left column shows the results of NICP without weighting the point correspondences. 

\begin{figure}[!t]
\centering
 \includegraphics[width=0.48\textwidth]{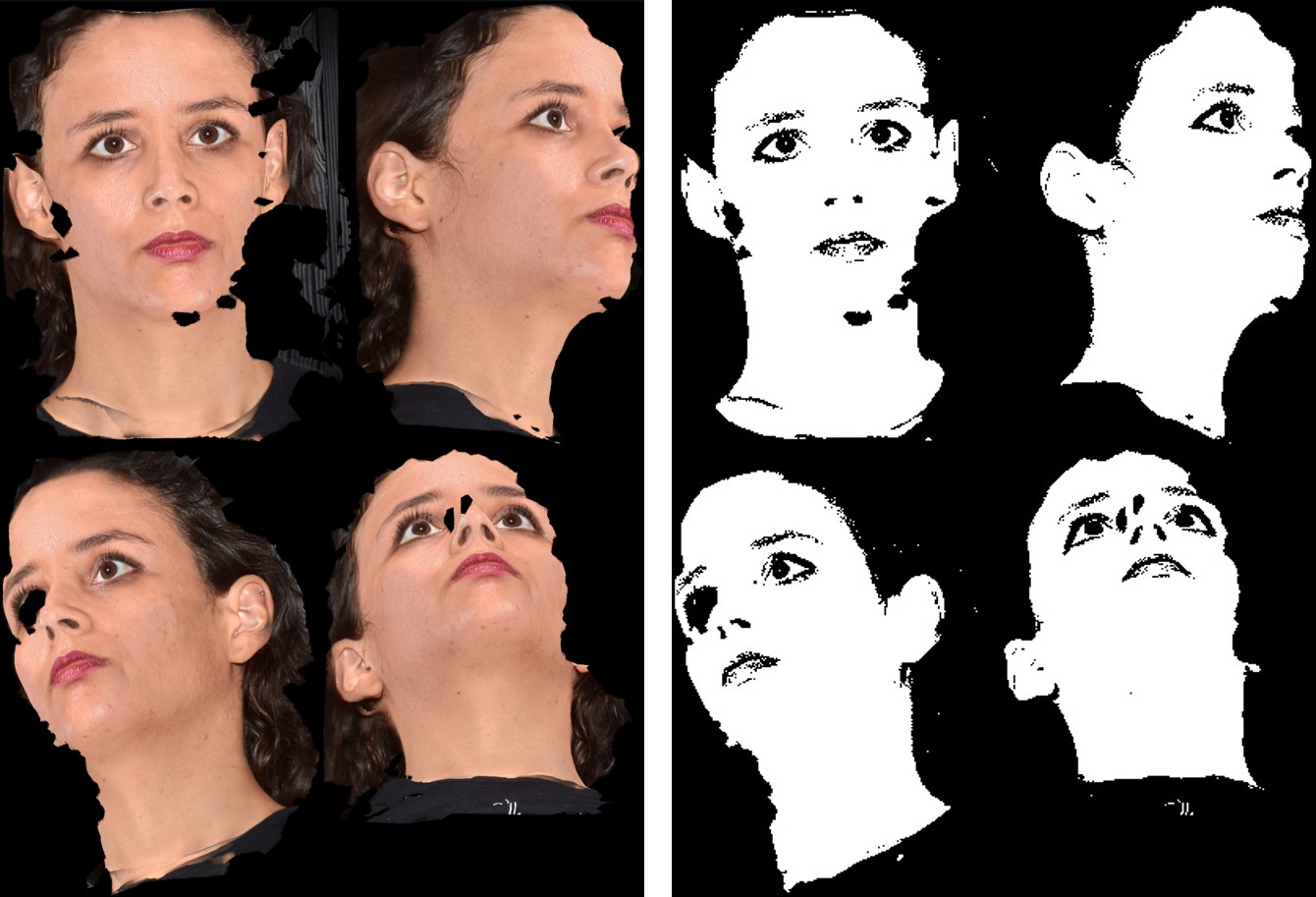}
\caption{Confidence map generated using a skin detection algorithm \cite{Kolkur2017}. The confidence map on the right can be used to weight the point correspondences in NICP.}
\label{fig:confidenceMap}
\end{figure}

The right column shows the result of NICP where the point correspondences are weighted according to a confidence map. The confidence map is shown in Figure \ref{fig:confidenceMap} and correspond to the skin area. As it is expected, the registration quality is better when we weight the point correspondences. For instance, the registration artifacts around the ears have disappeared. 

\section{DISCUSSION}\label{sec:discussion}

A first observation is that summing multiple kernels with varying variances does not  improve so much the results of registration. On these data, the Multiscale kernels do not demonstrate a real advantage (see Figure \ref{fig:regressionScores}) against single scale kernels. Indeed, the kernel shape is mostly determined by the smaller variances  and summing kernels has the effect of making the resulting distribution more heavy tailed.
%function tail heavier.  

Second,  B-spline  and Euclidean SE kernels tend to have similar regression energy profiles but their energy minima 
%(of regression energy) 
are not reached with the same $\sigma$ value. The reason is that  $\sigma$ defines the B-spline finite support for B-Splines and the standard deviation  for the SE kernel. Therefore we found smaller  $\sigma$ values for the Euclidean SE kernel. 

As expected, the geodesic SE kernel has a much lower registration energy when fitting facial expressions with mouth opening or eyes closing. For other facial expressions the geodesic SE kernel performs slightly better than others kernels. There is therefore a real interest in using geodesic distances. 
The registration energy has an impact on qualitative results. Figure \ref{fig:regressionResults} shows that the kernel with a geodesic metric is the only one that properly fit the mouth and the eyes (note that the color scale indicates the error in millimeters).

Furthermore, we remark that with small variances ($\sigma <20$) the error increases sharply. Indeed, as $\sigma$ values decrease, the approximation error increases since the number of basis vectors is fixed (and the eigenvalues are slowly decreasing with small $\alpha$.) Additionally, the energy remains low for a wide range of variance.
% we fixed  used in the low rank approximation, thus as we reduce the hyperparameter  .
The approximation error has a direct impact on the face expressiveness and thus on the energy. 

Finally, the standard deviation of errors through the various facial expressions is much lower for the geodesic SE kernel than for the Euclidean SE kernel (see Table ~\ref{tab:resultsPose}). This confirms that using the geodesic SE kernels makes the registration more robust with respect to varying facial expressions.

\section{CONCLUSIONS AND FUTURE WORKS}

\begin{figure}[tp]
 \centering
\begin{tabular}{lrrrrrr}
\toprule
{} & \multicolumn{2}{l}{Euclidean} & \multicolumn{2}{l}{Geodesic} & \multicolumn{2}{l}{Spline} \\
{} &      mean &   std &     mean &   std &   mean &   std \\
Pose    &           &       &          &       &        &       \\
\midrule
neutral  &      44.7 &   7.3 &     \textbf{41.3} &   7.7 &   45.4 &   7.9 \\
smile  &     244.6 &  39.0 &     \textbf{58.9} &  12.7 &  244.9 &  39.3 \\
stretch  &      62.8 &  13.5 &     \textbf{55.9} &  13.3 &   63.5 &  13.9 \\
anger  &      55.3 &  12.8 &    \textbf{51.5} &  12.7 &   55.9 &  13.8 \\
jaw left  &      56.5 &  13.8 &     \textbf{55.5} &  14.4 &   57.4 &  14.7 \\
jaw right  &      57.4 &  13.6 &     \textbf{53.8} &  13.5 &   57.2 &  14.7 \\
jaw forward  &      69.0 &  16.1 &     \textbf{51.6} &  13.1 &   71.7 &  16.5 \\
mouth left  &      71.6 &  19.2 &     \textbf{52.5} &  15.5 &   74.1 &  19.8 \\
mouth right   &      59.8 &  15.4 &     \textbf{50.3} &  14.4 &   61.2 &  16.1 \\
dimpler  &      62.7 &  12.9 &    \textbf{ 54.8} &  12.1 &   65.0 &  13.6 \\
chin raiser &      64.3 &  13.7 &    \textbf{54.5} &  12.8 &   65.4 &  14.5 \\
lip puckerer &      58.9 &  13.2 &    \textbf{53.7} &  12.9 &   61.5 &  13.6 \\
lip funneler &      58.4 &  14.1 &    \textbf{50.0} &  14.0 &   60.4 &  14.8 \\
sadness &      66.6 &  13.2 &     \textbf{55.0} &  12.5 &   70.2 &  14.6 \\
lip roll &      90.4 &  18.1 &    \textbf{61.0} &  12.6 &   96.5 &  18.5 \\
grin &      58.7 &  13.2 &     \textbf{53.6} &  12.5 &   60.4 &  13.8 \\
blowing &      65.3 &  13.0 &    \textbf{64.6} &  12.0 &   74.1 &  12.8 \\
eye closed &     154.9 &  26.3 &     \textbf{63.0} &  13.7 &  155.5 &  26.2 \\
brow raiser &      58.8 &  16.0 &     \textbf{50.9} &  15.8 &   59.8 &  16.5 \\
brow lower &      77.4 &  12.7 &     \textbf{53.8} &  13.4 &   76.0 &  13.2 \\
\bottomrule
\end{tabular}
\caption{This table shows the average and standard deviation of registration energy $R$ results for the 20 facial expressions of the FaceWarehouse dataset using different kernels. The results have been computed with single scale kernels and optimized hyperparameters. Best results for each expression are shown in bold font.}
\label{tab:resultsPose}
\end{figure}

\subsection{Conclusion}
%ABl: Pourquoi peut-on dire que ça "simplifie le recalage"? Qu'est-ce qui le rend plus simple que d'autres noyaux?
Our work simplifies the registration of a template mesh towards 3D face scans, which is still the standard approach for shape modeling. To do so, we extended the GPMMs framework by proposing a new kernel that takes into account the geodesic distance between mesh vertices to create a deformation prior. We demonstrated that the GPMMs framework and NICP can now be used to fit a template mesh to 3D face scans with different facial expressions. We also learned the kernel's hyperparameters so that they can be reused for the registration of additional face mesh datasets. We tested the proposed geodesic kernel and the learned hyperparameters for the purpose of registration on the whole FaceWarehouse dataset.
We have shown that using geodesic SE kernels significantly improves the quality of the registration for varying facial expressions in comparison to other kernels based on Euclidean distances.

\subsection{Future Work}

There is certainly a theoretical work to be done on the positive definiteness of geodesic SE kernels. Indeed, it has been shown that in the general case the positive definiteness does not hold and yet geodesic kernels are specifically useful for modeling facial expressions. There is a background work to identify the cases in which we can rigorously use geodesic kernels.

One aspect that has not been studied here is that B-spline kernels generate sparse Gram matrices, which is potentially useful for further algorithmic optimizations.

Finally, one may use Laplacian  kernels (i.e. the case $q=1$ in equation (1) of \cite{Feragen2015})  instead of Gaussian kernels. Indeed, Laplacian kernels are heavy tailed distributions which may replace advantageously the sum of multiple Gaussian kernels.
%make the kernel looking like an heavy tailed distribution has the laplacian distribution may be. 

%%%%%%%%%%%%%%%%%%%%%%%%%%%%%%%%%%%%%%%%%%%%%%%%%%%%%%%%%%%%%%%%%%%%%%%%%%%%%%%%
\section*{Acknowledgements}
This work was supported by the ANRT Cifre contract 2019/0101 with QuantifiCare \& INRIA and  by the French government through the 3IA Côte d'Azur Investments ANR-19-P3IA-0002 managed by the National Research Agency.

%We would like to thank the Graphics and Parallel Systems Lab, of Zhejiang University for providing the FaceWarehouse dataset and the CITE LAB of Nanjing University for providing the FaceScape dataset. 
%We also thanks QuantifiCare and the ANRT (Association nationale de la recherche et de la technologie.) for funding this research.     

%The authors gratefully acknowledge the contribution of reviewers' comments, etc. (if %desired). Put sponsor acknowledgments in the unnumbered footnote on the first page.
%
%
%%%%%%%%%%%%%%%%%%%%%%%%%%%%%%%%%%%%%%%%%%%%%%%%%%%%%%%%%%%%%%%%%%%%%%%%%%%%%%%%%
%
%References are important to the reader; therefore, each citation must be complete and %correct. If at all possible, references should be commonly available publications.

{\small
\bibliographystyle{ieee}
\bibliography{egbib}
}

\end{document}